\documentclass{article}
\usepackage{authblk}
\usepackage[utf8]{inputenc}
\usepackage{main}
\usepackage{microtype}
\usepackage{subcaption}
\usepackage{graphicx}
\usepackage{times}
\usepackage{latexsym}
\usepackage{amsmath}
\usepackage{float}
\usepackage{footnote}
\usepackage{enumitem}
\usepackage{bm}
\usepackage{arydshln}
\usepackage{booktabs}
\usepackage{array}     
\usepackage{multicol}
\usepackage{multirow}
\usepackage{color}
\usepackage{xcolor}     
\usepackage{colortbl}
\usepackage{bbding}
\usepackage{makecell}
\usepackage{mathtools}
\usepackage{imakeidx}
\usepackage{longtable}
\usepackage{wrapfig}
\usepackage{algorithmic}
\usepackage{rotating}
\makeindex
\usepackage{arydshln}
\usepackage{lipsum}
\usepackage{natbib}
\usepackage[toc]{multitoc}
\usepackage[edges]{forest}
\usepackage[normalem]{ulem}
\definecolor{mydarkblue}{rgb}{0,0.08,0.45}
\usepackage[colorlinks=true,linkcolor=mydarkblue,citecolor=mydarkblue,filecolor=mydarkblue,urlcolor=mydarkblue]{hyperref}
\usepackage{CJKutf8}
\usepackage{awesomebox} 
\usepackage{bbding}
\usepackage[most]{tcolorbox}
\usepackage{booktabs}
\usepackage{geometry}
\definecolor{wkblue}{rgb}{0.2, 0.3, 0.6}
\definecolor{meta-color}{rgb}{0.5, 0.5, 0.5}
\usepackage{amsmath}
\usepackage{enumitem}
\usepackage{lscape} 
\usepackage{booktabs}
\usepackage{algorithm}   
\usepackage{setspace}
\usepackage{comment}

\usepackage{algorithmic}
\usepackage{tabularx,booktabs}
\usepackage{makecell}

\usepackage{amssymb}
\usepackage{amsfonts}

\usepackage{booktabs} 
\usepackage{geometry} 

\usepackage{multirow}

\usepackage[tikz]{bclogo}
\usepackage[framemethod=tikz]{mdframed}
\definecolor{bgblue}{RGB}{245,243,253}
\definecolor{ttblue}{RGB}{91,194,224}

\usepackage{pgfplots}
\usepackage{pgfplotstable}

\usepackage{wrapfig}
\usepackage{graphicx}

\mdfdefinestyle{mystyle}{%
  rightline=true,
  innerleftmargin=10,
  innerrightmargin=10,
  outerlinewidth=3pt,
  topline=false,
  rightline=true,
  bottomline=false,
  skipabove=\topsep,
  skipbelow=\topsep
}

\newtcolorbox{myboxi}[1][]{
  breakable,
  title=#1,
  colback=red!5,
  colbacktitle=red!5,
  coltitle=black,
  fonttitle=\bfseries,
  bottomrule=0pt,
  toprule=0pt,
  leftrule=2pt,
  rightrule=2pt,
  titlerule=0pt,
  arc=0pt,
  outer arc=0pt,
  colframe=red,
}

\newtcolorbox{myboxnote}[1][]{
  breakable,
  title=#1,
  colback=orange!0,
  colbacktitle=orange!0,
  coltitle=black,
  fonttitle=\bfseries,
  bottomrule=0pt,
  toprule=0pt,
  leftrule=2pt,
  rightrule=2pt,
  titlerule=0pt,
  arc=0pt,
  outer arc=0pt,
  colframe=orange,
}

\newtcolorbox{myboxii}[1][]{
  breakable,
  freelance,
  title=#1,
  colback=white,
  colbacktitle=white,
  coltitle=black,
  fonttitle=\bfseries,
  bottomrule=0pt,
  boxrule=0pt,
  colframe=white,
  overlay unbroken and first={
  \draw[red!75!black,line width=3pt]
    ([xshift=5pt]frame.north west) -- 
    (frame.north west) -- 
    (frame.south west);
  \draw[red!75!black,line width=3pt]
    ([xshift=-5pt]frame.north east) -- 
    (frame.north east) -- 
    (frame.south east);
  },
  overlay unbroken app={
  \draw[red!75!black,line width=3pt,line cap=rect]
    (frame.south west) -- 
    ([xshift=5pt]frame.south west);
  \draw[red!75!black,line width=3pt,line cap=rect]
    (frame.south east) -- 
    ([xshift=-5pt]frame.south east);
  },
  overlay middle and last={
  \draw[red!75!black,line width=3pt]
    (frame.north west) -- 
    (frame.south west);
  \draw[red!75!black,line width=3pt]
    (frame.north east) -- 
    (frame.south east);
  },
  overlay last app={
  \draw[red!75!black,line width=3pt,line cap=rect]
    (frame.south west) --
    ([xshift=5pt]frame.south west);
  \draw[red!75!black,line width=3pt,line cap=rect]
    (frame.south east) --
    ([xshift=-5pt]frame.south east);
  },
}

\usepackage{fontawesome5}
\usepackage{fancyhdr} 
\usepackage{blindtext} 
\usepackage{makecell}

\DeclareCaptionFont{black}{\color{black}}

\definecolor{myblue}{rgb}{0.9, 0.1, 0.94}
\definecolor{mygreen}{rgb}{0.64, 0.56, 0.88}
\definecolor{myyellow}{rgb}{0.68, 0.6, 0.1}
\definecolor{fancygreen}{rgb}{0.33, 0.68, 0.20}
\definecolor{salmon}{rgb}{0.94, 0.52, 0.49}
\definecolor{tablegreen}{rgb}{0.82, 0.94, 0.75}
\definecolor{tableblue}{rgb}{0.81, 0.90, 0.94}
\definecolor{tablered}{rgb}{0.97, 0.85, 0.85}
\definecolor{tableorange}{rgb}{0.96, 0.85, 0.81}

\newenvironment{itemize*}%
 {\leftmargini=10pt\begin{itemize}%
  \setlength{\itemsep}{0pt}%
  \setlength{\parskip}{0pt}%
  }%
 {\end{itemize}}
\newenvironment{enumerate*}%
 {\begin{enumerate}%
  \setlength{\itemsep}{0pt}%
  \setlength{\parskip}{0pt}}%
 {\end{enumerate}}

\usepackage{xcolor}
\usepackage{listings}  

\newcommand\JSONnumbervaluestyle{\color{blue}}
\newcommand\JSONstringvaluestyle{\color{red}}

\newif\ifcolonfoundonthisline

\makeatletter

\lstdefinestyle{json}
{
  showstringspaces    = false,
  keywords            = {false,true},
  alsoletter          = 0123456789.,
  morestring          = [s]{"}{"},
  stringstyle         = \ifcolonfoundonthisline\JSONstringvaluestyle\fi,
  MoreSelectCharTable =%
    \lst@DefSaveDef{`:}\colon@json{\processColon@json},
  basicstyle          = \ttfamily,
  keywordstyle        = \ttfamily\bfseries,
}

\newcommand\processColon@json{%
  \colon@json%
  \ifnum\lst@mode=\lst@Pmode%
    \global\colonfoundonthislinetrue%
  \fi
}

\lst@AddToHook{Output}{%
  \ifcolonfoundonthisline%
    \ifnum\lst@mode=\lst@Pmode%
      \def\lst@thestyle{\JSONnumbervaluestyle}%
    \fi
  \fi
  \lsthk@DetectKeywords%
}

\lst@AddToHook{EOL}%
  {\global\colonfoundonthislinefalse}

\makeatother

\usepackage{etoolbox}
\usepackage{natbib}
\usepackage{url}
\newcounter{bibcount}
\makeatletter
\patchcmd{\@lbibitem}{\item[}{\item[\hfil\stepcounter{bibcount}{[\thebibcount]}}{}{}
\renewcommand\NAT@bibsetup%
  [1]{\setlength{\leftmargin}{\bibhang}\setlength{\itemindent}{-\parindent}%
      \setlength{\itemsep}{\bibsep}\setlength{\parsep}{\z@}}
\makeatother

\definecolor{mybrown}{RGB}{128,64,0}

\definecolor{titlecolor}{HTML}{4c9cff}
\definecolor{coolblue3}{rgb}{0.91, 0.94, 0.98}

\begin{document}

\title{LiveTalk: Real-Time Multimodal Interactive Video Diffusion via Improved On-Policy Distillation}

\author{
Ethan Chern\textsuperscript{*} \quad
Zhulin Hu\textsuperscript{*} \quad
Bohao Tang\textsuperscript{*} \quad
Jiadi Su\textsuperscript{*} \quad
Steffi Chern \quad
Zhijie Deng\textsuperscript{$\dagger$} \quad
Pengfei Liu\textsuperscript{$\dagger$}\textsuperscript{$\ddagger$}
\\
SII \quad SJTU \quad GAIR
} 

\maketitle
\pagestyle{fancy}
\thispagestyle{fancy}
\fancyhead{}


\rhead{%
  \raisebox{-0.3cm}{\includegraphics[height=0.7cm]{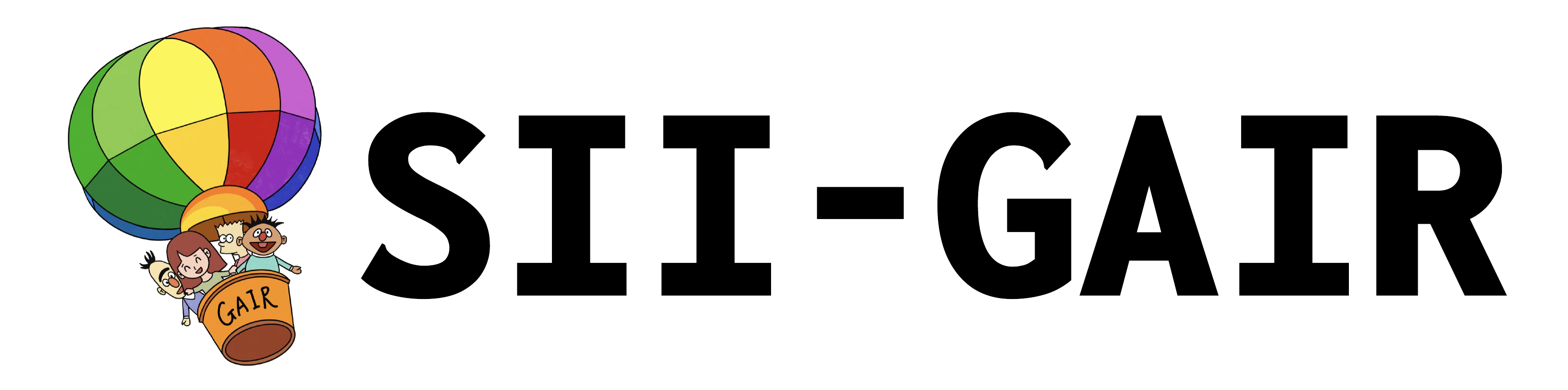}}%
}
\renewcommand{\headrulewidth}{0pt}


\renewcommand{\thefootnote}{}
\footnotetext{* Equal contribution.}
\footnotetext{$\dagger$ Co-advising.}
\footnotetext{$\ddagger$ Corresponding author.}
\vspace{-20pt}

{\centering
%
\href{https://github.com/GAIR-NLP/LiveTalk}{\textcolor{black}\faGithub\ Code}
\quad \href{https://huggingface.co/GAIR/LiveTalk-1.3B-V0.1}{\raisebox{-.15em}{\includegraphics[height=1em]{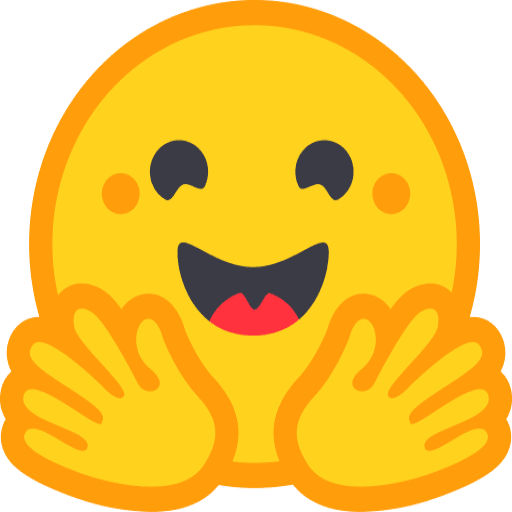}}\ Models}
\par}

\vspace{20pt}

\begin{abstract}
Real-time video generation via diffusion is essential for building general-purpose multimodal interactive AI systems. 
However, the simultaneous denoising of all video frames with bidirectional attention via an iterative process in diffusion models prevents real-time interaction. 
While existing distillation methods can make the model autoregressive and reduce sampling steps to mitigate this, they focus primarily on text-to-video generation, leaving the human-AI interaction unnatural and less efficient. 
This paper targets real-time interactive video diffusion conditioned on a multimodal context, including text, image, and audio, to bridge the gap. Given the observation that the leading on-policy distillation approach Self Forcing encounters challenges (visual artifacts like flickering, black frames, and quality degradation) with multimodal conditioning, we investigate an improved distillation recipe with emphasis on the quality of condition inputs as well as the initialization and schedule for the on-policy optimization. On benchmarks for multimodal-conditioned (audio, image, and text) avatar video generation including HDTF~\cite{zhang2021flow}, AVSpeech~\cite{ephrat2018looking}, and CelebV-HQ~\cite{zhu2022celebv}, our distilled model matches the visual quality of the full-step, bidirectional baselines of similar or larger size with 20× less inference cost and latency. Further, we integrate our model with audio language models and long-form video inference technique Anchor-Heavy Identity Sinks to build LiveTalk, a real-time multimodal interactive avatar system. System-level evaluation on our curated multi-turn interaction benchmark shows LiveTalk outperforms state-of-the-art models (Sora2, Veo3) in multi-turn video coherence and content quality, while reducing response latency from 1 to 2 minutes to real-time generation, enabling seamless human-AI multimodal interaction.

\begin{figure*}[h]
    \centering
    \includegraphics[page=1, height=4cm, width=\textwidth]{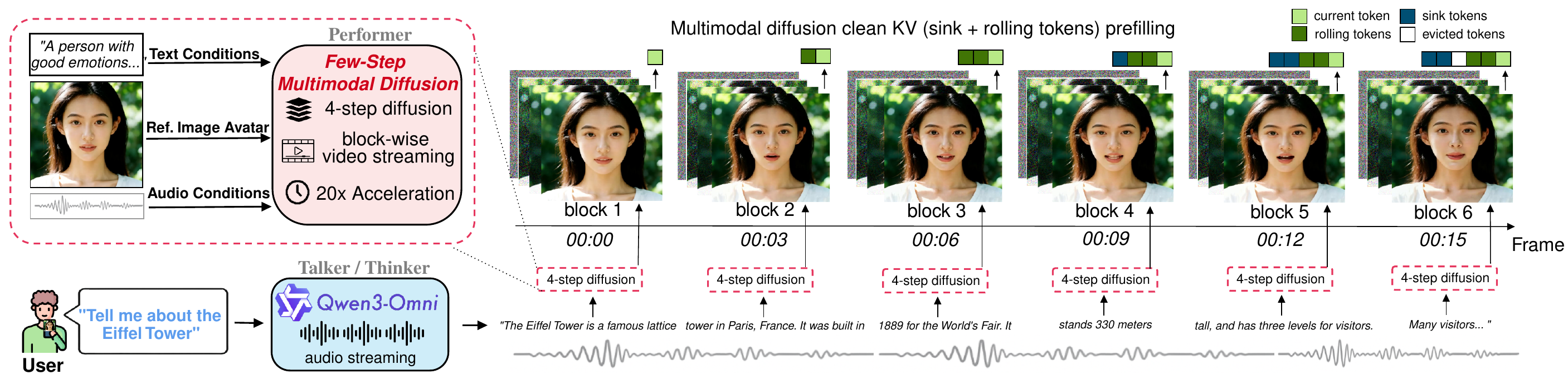}
    \vspace{-4mm}
    \caption{\textbf{Overview of the LiveTalk system.} Given a user audio/text query, Qwen3-Omni~\cite{xu2025qwen3} processes the query and generates streaming audio responses in real-time. Our few-step multimodal diffusion model takes the streaming audio along with the reference image avatar and text conditions to generate synchronized video responses through block-wise AR generation.
    Each block (3 latent frames) undergoes 4-step diffusion, achieving 20× acceleration over the baseline (See Tab.~\ref{tab:face}, Ours vs. OmniAvatar-1.3B~\cite{gan2025omniavatar}). To support long-horizon streaming with sub-second latency, we perform clean KV prefilling across blocks using a \emph{sink+rolling} token cache: persistent sink tokens retain global context, rolling tokens carry recent history.}
    \vspace{-4mm}
    \label{fig:system_overview}
\end{figure*}

\end{abstract}

\pagestyle{fancy}
\lhead{\rightmark}
\renewcommand{\headrulewidth}{0.7pt}
\setlength{\headsep}{5mm}

\clearpage

\newpage

\renewcommand{\thefootnote}{\arabic{footnote}}
\setcounter{footnote}{0}  

\section{Introduction}
\label{sec:intro}
Diffusion transformers (DiTs)~\cite{peebles2023scalable, brooks2024video, wan2025wan, chen2025hunyuanvideo} have enabled appealing visual fidelity for video generation. 
The sampling involves denoising all video frames simultaneously using bidirectional attention via an iterative process. 
It can be costly, e.g., state-of-the-art models such as Veo3~\cite{deepmind_veo3_2025} and Sora2~\cite{openai_sora2_2025} require 1 to 2 minutes to generate a single 5 to 10 second clip, incurring prohibitive inference expense and creating fundamental barriers to real-time applications. 

As a result, there is increasing interest in converting pretrained, bidirectional, many-step video diffusion models into causal, few-step autoregressive (AR) ones via distillation techniques~\cite{lu2025onpolicydistillation, agarwal2024policy, yin2024one, huang2025self, liu2025rolling}. The resultant models enable low-latency streaming generation. However, systematic investigation of complex multimodal conditioning for distillation remains largely underexplored, particularly the simultaneous integration of text, images, and audio for interactive avatar generation. Such natural conditioning 
is critical for building general-purpose interactive AI systems that can not only \textit{understand} across modalities but also \textit{express themselves visually} for natural human-AI interaction. 

This paper aims to establish a real-time interactive video diffusion model conditioned on a multimodal context. 
We first perform an in-depth investigation based on the leading on-policy distillation approach, Self Forcing~\cite{huang2025self}, which involves an initialization stage based on ODE trajectory distillation~\cite{song2023consistency, yin2025slow, gu2023boot, berthelot2023tract} to obtain a few-step causal student model with block-wise causal attentions, as well as a distribution matching distillation (DMD)~\cite{yin2024one, yin2024improved, yin2025slow} stage to minimize \textit{exposure bias}~\cite{ning2023elucidating, schmidt2019generalization, huang2025self} from causality based on on-policy rollouts. We observe Self Forcing can result in extensive visual artifacts, e.g., flickering effects (see row 1 of Fig.~\ref{fig:ablation_visual}) for the multimodal setting, with related issue reported in the community~\cite{chen2025causvid_issue51}. We speculate that the issues may stem from the complex interplay among the inherent multiple components in DMD, which renders the optimization unstable and fragile, particularly under complex multimodal conditions.

To address this, we investigate an improved distillation recipe with emphasis on the quality of the conditions as well as the initialization and schedule for a stable on-policy optimization.
Concretely, we advocate (1) \textbf{refining multimodal conditions for distillation}, e.g., making the condition image high-quality and the text prompts motion-focused; 
(2) \textbf{training ODE initialization to convergence} before applying on-policy DMD training; and (3) \textbf{maximizing learning within DMD's limited learning window} through aggressive learning rates and tuned classifier-guidance (CFG)~\cite{ho2022classifier} scales. 
We validate these by distilling a multimodal variant of Wan2.1~\cite{wan2025wan} (i.e., OmniAvatar~\cite{gan2025omniavatar}) and benchmarking on diverse multimodal-driven avatar generation datasets, including HDTF~\cite{zhang2021flow}, AVSpeech~\cite{ephrat2018looking}, and CelebV-HQ~\cite{zhu2022celebv}. Our distilled model even surpasses some 5B and 14B bidirectional, many-step baselines in multiple aspects. The distillation process dramatically improves inference efficiency: achieving 24.82 FPS (20× speedup compared to the vanilla model), and reducing first-frame latency to subsecond (200× speedup), opening the door for real-time interactive communication. 

Based on the distilled model, we build a real-time multimodal avatar system, LiveTalk, that enables seamless interaction between humans and AI. 
The system (Fig.~\ref{fig:system_overview}) leverages existing audio language models~\cite{xu2025qwen3} for reasoning and speech, 
while our model renders talking avatars in real-time with high visual fidelity (Fig.~\ref{fig:system_overview}).  To preserve speaker identity in long-horizon video streaming, we introduce a training-free technique, the Anchor-Heavy Identity Sinks (AHIS), which successfully keeps the generated speaker visually undistorted on a time scale of minutes. We also curate a new multi-turn interaction benchmark for this new form of real-time multimodal talking avatar systems. 
Evaluations against state-of-the-art models Veo3~\cite{deepmind_veo3_2025} and Sora2~\cite{openai_sora2_2025} show that our system substantially outperforms them in multi-round coherence and content quality, while reducing response latency from minutes to real-time generation, enabling truly interactive communication. 

In summary, our contributions are: 
\begin{itemize}
    \item \textbf{Actionable distillation framework for multimodal video diffusion.} 
    We establish a systematic recipe for training real-time multimodal interactive video models conditioned on text, image, and audio. We identify three key improvements for stable on-policy distillation under complex multimodal conditions: curated multimodal conditioning, converged ODE initialization, and aggressive optimization schedule. Our ablation study (Tab.~\ref{tab:ablation-four-factors}) demonstrates these collectively deliver significant quality improvements across perceptual metrics, audio-visual synchronization, and aesthetic quality.

    \item \textbf{Real-time multimodal video generation with 20× speedup.} Our distilled 4-step model matches or exceeds bidirectional diffusion baselines while reducing inference cost by over 20×. Our 1.3B model matches or surpasses the 1.3B bidirectional variant (OmniAvatar-1.3B~\cite{gan2025omniavatar}) and larger baselines, including Hallo3~\cite{cui2024hallo3}, FantasyTalking~\cite{wang2025fantasytalking}, and AniPortrait~\cite{wei2024aniportrait} across quality metrics, while achieving real-time generation at 24.82 FPS on a single GPU.

    \item \textbf{Complete real-time multimodal interactive avatar system.} We build LiveTalk and propose a benchmark for evaluating multi-turn interaction quality of multimodal interactive avatar system. Our system outperforms Veo3~\cite{deepmind_veo3_2025} and Sora2~\cite{openai_sora2_2025} in multi-video coherence and content quality metrics, while achieving sub-second response, enabling seamless human-AI interaction.
\end{itemize}

\section{Related Work}

\subsection{Multimodal Video Diffusion}

Modern video synthesis has evolved beyond text control to incorporate richer conditioning signals such as images and audio, enhancing controllability and versatility~\cite{wan2025wan, kong2024hunyuanvideo, chen2025hunyuanvideo, gan2025omniavatar}. These multimodal capabilities enable applications including image-guided video editing~\cite{wan2025wan, kong2024hunyuanvideo} and multimodal-driven virtual avatars~\cite{gan2025omniavatar, chen2025hunyuanvideo}. Extending video diffusion to handle complex multimodal conditions presents challenges including architectural designs~\cite{ju2025fulldit, lin2025stiv, hu2025hunyuancustom}, training overheads~\cite{wan2025wan, kong2024hunyuanvideo}, and cross-modal alignment~\cite{li2024animate, gan2025omniavatar, chen2025hunyuanvideo}. However, existing multimodal video diffusion predominantly adopts pure diffusion paradigms requiring many-step iterative denoising across entire sequences. While achieving high visual quality, their significant inference cost and latency render them impractical for real-time applications. While recent work~\cite{low2025talkingmachines} has begun exploring real-time audio-driven models, comprehensive recipes for training stability, exposure bias mitigation, and rigorous benchmarking remain largely underexplored.

\subsection{Real-Time Video Diffusion}

Real-time video diffusion adopts hybrid modeling combining AR and diffusion components: AR enables streaming generation without future frame dependencies, while diffusion ensures high visual fidelity~\cite{teng2025magi, bruce2024genie, liu2025rolling, yang2025longlive}. However, AR video diffusion faces \textit{exposure bias}~\cite{ning2023elucidating, schmidt2019generalization, huang2025self} from error accumulation during autoregression. Post-training strategies address this through distilling guidance from bidirectional models~\cite{yin2025slow}, increasing robustness to imperfect frames~\cite{chen2024diffusion}, and mitigating train-test gaps via self-generated rollouts~\cite{huang2025self}. However, these techniques have primarily been explored for text-to-video and remain largely unexamined for multimodal video diffusion. We bridge this gap by systematically investigating and improving on-policy distillation for multimodal conditions (text, image, audio), emphasizing condition quality, initialization, and optimization dynamics, and establishing a comprehensive multi-turn interaction benchmark to guide future development for real-time avatar systems.

\section{Improved Distillation for Real-Time Multimodal Interactive Video Diffusion}
\label{sec:sec3}

This section briefly reviews Self Forcing~\cite{huang2025self}, a dominant distillation method for constructing real-time video diffusion, discusses its limitations for handling video diffusion models with multimodal conditioning, and elaborates on our improved distillation strategies.

\subsection{Preliminary: Self Forcing}

There has been ongoing interest in transferring bidirectional, many-step, pure diffusion models~\cite{wan2025wan, chen2025hunyuanvideo} into causal, few-step AR ones via distillation-based post-training approaches~\cite{chen2024diffusion, huang2025self, yin2025slow}. Self Forcing~\cite{huang2025self} is one of the most effective approaches, which follows a two-stage procedure.
First, an ODE initialization~\cite{song2023consistency, yin2025slow, gu2023boot, berthelot2023tract} stage is performed to obtain a few-step causal student model with block-wise causal attentions. 
Then, it conducts on-policy distillation~\cite{lu2025onpolicydistillation, agarwal2024policy} with self-generated rollouts~\cite{huang2025self} using the distribution matching distillation (DMD) principle~\cite{yin2024one, yin2024improved, yin2025slow}, which helps minimize the \textit{exposure bias}~\cite{ning2023elucidating, schmidt2019generalization, huang2025self} of the student model. 

Concretely, let $\mathbf{x}_0$ denote the latent corresponding to video frames yielded by a compression variational auto-encoder (VAE)~\cite{kingma2013auto, rombach2022high}. 
Letting $\mathbf{c}$ denote the conditioning for generation. Self Forcing aims at distilling a vanilla teacher model into a student one $g_\phi$ that can generate in a block-by-block manner, where each block consists of multiple latent frames (e.g., 3 in our case) in real-time. 

\noindent \textbf{ODE Initialization}
first performs trajectory distillation. 
It subsamples $k = 4$ timesteps from the teacher's $N = 48$ step denoising trajectory $\{\mathbf{x}_{t_j}\}_{j=0}^{N}$, based on which the causal student is asked to predict clean $\mathbf{x}_0$:
\begin{equation}
      \mathcal{L}_{\text{ODE}} = \mathbb{E}_{t \sim \{t_{iN/k}\}_{i=0}^{k-1}}
      \left[\sum_{b} \left\| g_\phi(\mathbf{x}_{t}^b, t, \mathbf{c}) - \mathbf{x}_{0}^b \right\|_2^2\right],
\end{equation}
with the superscript $b$ denoting the $b$-th block. 

\noindent \textbf{Distribution Matching Distillation (DMD)} then addresses the exposure bias~\cite{yin2024one, yin2024improved, yin2025slow} of the model after ODE initialization due to the training under teacher-forced trajectories. 
The algorithmic implementation introduces two extra models:  
a frozen teacher score network $s_\theta$~\cite{song2020score,yin2024one,yin2024improved} and a trainable critic $s_\psi$, and the training alternates between updating the generator $g_\phi$ and the critic $s_\psi$.
Specifically, the gradient for updating the student model is 
\begin{equation}
    -\mathbb{E}_{\tau, \hat{\mathbf{x}}_0, \mathbf{x}_\tau}\left[\left(s_{\theta}(\mathbf{x}_\tau, \tau, \mathbf{c}) - s_{\psi}(\mathbf{x}_\tau, \tau, \mathbf{c})\right) \frac{\partial \hat{\mathbf{x}}_0}{\partial \phi}\right],
\end{equation}
where $\hat{\mathbf{x}}_0 = g_\phi(\mathbf{z}, 0, \mathbf{c})$ with $\mathbf{z} \sim \mathcal{N}(0, I)$ and $\mathbf{x}_\tau$ comes from adding $\tau$-time noise to $\hat{\mathbf{x}}_0$. 
Note that, for multimodal conditioning, $s_\theta$ can be estimated with the classifier-free guidance~\cite{ho2022classifier} strategy using separate scales for various conditions. 
The critic $s_{\psi}$ learns to track the evolving distribution of the generator by 
minimizing the standard diffusion denoising objective:
\begin{equation}
    \mathcal{L}_{\text{critic}} = \mathbb{E}_{\tau} \left[\left\| s_{\psi}(\mathbf{x}_\tau, \tau, \mathbf{c}) - \hat{\mathbf{x}}_0 \right\|_2^2\right].
\end{equation}

\subsection{Issues on Existing Distillation Recipe}
\begin{figure*}[htbp]
    \centering
    \includegraphics[page=1, width=0.97\textwidth]{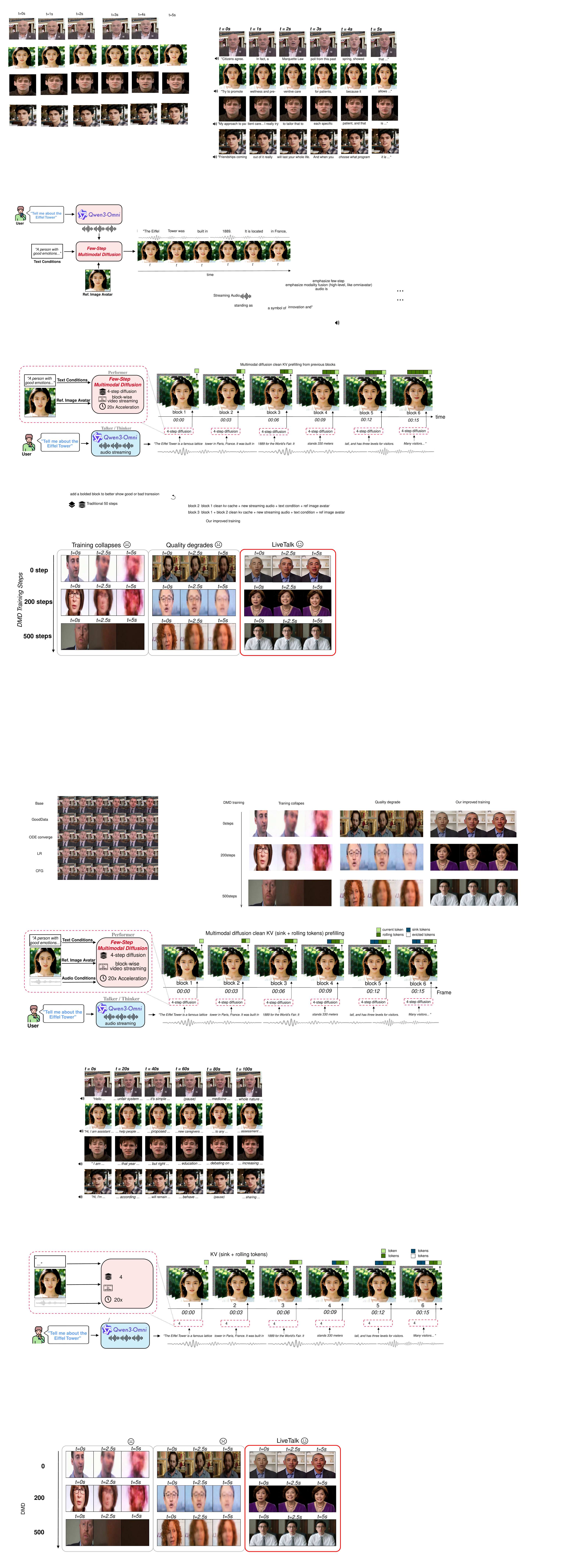}
    \vspace{-2mm}
    \caption{\textbf{Degraded training performance with Self Forcing DMD.} Left and middle columns show failure cases from Self Forcing DMD training, exhibiting quality degradation. Right column shows stable results from our method.}
    \vspace{-4mm}
    \label{fig:bad_case}
\end{figure*}

\label{subsec:issues}

When naively applying on-policy distillation with default settings from Self Forcing~\cite{huang2025self} to distill multimodal video diffusion models, we encounter significant training instabilities that manifest as visual artifacts (Fig.~\ref{fig:bad_case}, left and middle).
This challenge likely stems from the complex interplay in the DMD training, where the critic score network $s_\psi$ learns to denoise the noised generator rollout $\mathbf{x}_\tau$ in the critic training stage, creating a delicate interdependency. When the generator output degrades catastrophically, the critic score network $s_\psi$ receives corrupted training signals, resulting in inaccurate gradient estimates that further degrade the generator, potentially triggering mode collapse~\cite{ge2025senseflowscalingdistributionmatching}. While Self Forcing demonstrates robustness under text conditioning, multimodal conditions introduce additional complexity that could amplify instability~\cite{chen2025causvid_issue51}. Through investigative studies, we motivate three critical factors that contribute to the instability:

\noindent \textbf{Data Quality Issues.}
In our initial trials, we selected 2000 multimodal conditions (reference image and audio) from each of the Hallo3 and HDTF datasets and applied distillation with default settings of Self Forcing. However, the DMD training collapsed after several hundred iterations, with outputs degrading to black images (Fig.~\ref{fig:bad_case}, left). Through investigation, we discovered that the quality of the reference image condition critically influences distillation stability. Specifically, existing datasets often contain lower-quality images with artifacts (such as Hallo3's overall low image quality and HDTF's facial blurriness) which lead to imperfect generator rollouts during distillation. These imperfect rollouts in turn produce corrupted training signals that destabilize the learning process. To address this issue, we filtered distillation training samples using brightness and quality metrics, which yielded slight improvements in training stability (Fig.~\ref{fig:bad_case}, middle). However, we still observed blurriness emerging after a few hundred training steps, suggesting that additional factors contribute to distillation instability, as we discuss below.

\noindent \textbf{Insufficient ODE Initialization.} 
We observe that insufficient ODE initialization creates a weak foundation that leads to instability during subsequent on-policy DMD training. Unlike text-to-video distillation~\cite{huang2025self} where even ODE checkpoints generating low-quality videos~\cite{selfforcing_issue50} can lead to successful distillation, multimodal-conditioned distillation exhibits markedly different behavior. Insufficient ODE training manifests as severity-dependent DMD failures: collapse in extreme cases, and performance plateaus with blurry artifacts in moderate cases. This suggests that distilling multimodal video diffusion requires a more robust starting point to facilitate the critic-generator interdependency to function stably.

\noindent \textbf{Limited Learning Window.}
We observe that the effective learning window for multimodal-conditioned DMD training is considerably short, with the model reaching peak performance within a few hundred steps before degrading. In contrast to text-to-video distillation where prior work reports convergence in 90 minutes, multimodal conditioning exhibits a peak-then-degrade pattern~\cite{chen2025causvid_issue51}. With standard learning rates and guidance scales, the model fails to sufficiently learn optimal multimodal alignment (particularly audio-visual synchronization) before degradation occurs, leaving considerable performance on the table.

These three factors interact to create a relatively fragile on-policy distillation training procedure that careful treatment beyond what is required for text-to-video distillation.
\subsection{Improvements on Existing Distillation Recipe}
\label{subsec:improvements}

To address the challenges identified in Section~\ref{subsec:issues}, we propose three key improvements to the on-policy distillation procedure for multimodal diffusion models. We demonstrate the effectiveness of each component through ablations in Tab.~\ref{tab:ablation-four-factors} and Fig.~\ref{fig:ablation_visual}.

\noindent \textbf{Refining Multimodal Conditions for Distillation.}
Instead of using existing dataset directly, we meticulously curate high-quality multimodal conditions $\mathbf{c} = \{\mathbf{c}_{\text{text}}, \mathbf{c}_{\text{img}}, \mathbf{c}_{\text{audio}}\}$ to provide clean training signals. Note that we mainly consider improving $\mathbf{c}_{\text{img}}$ and $\mathbf{c}_{\text{text}}$ in this paper. We adopt targeted curation strategies for different dataset. For Hallo3, characterized by overall low image quality, we employ Qwen-Image~\cite{wu2025qwen} to generate semantically consistent yet high-quality reference frames $\mathbf{c}_{\text{img}}$. For HDTF, which primarily suffers from facial blurriness, we apply super-resolution~\cite{zhou2022robustblindfacerestoration} to obtain clear facial details. Furthermore, we utilize Qwen2.5-VL-72B~\cite{bai2025qwen2} to refine text prompts $\mathbf{c}_{\text{text}}$, emphasizing dynamic motion and facial expressions to enrich temporal and semantic information. 

\noindent \textbf{Training ODE Initialization to Convergence.}
To establish a robust starting point for on-policy distillation, we train ODE initialization to full convergence using an extended training schedule, ensuring that the student model has thoroughly learned to denoise across all timesteps before transitioning to DMD. This convergent ODE checkpoint provides a strong foundation that stabilizes the delicate critic-generator interdependency during subsequent on-policy training.

\noindent \textbf{Maximizing Learning Within the Limited Window of On-Policy Distillation.}
To maximize learning before degradation occurs, we employ an aggressive learning rate schedule (2$\times$ baseline) that accelerates convergence within the limited effective learning window. We also apply higher CFG guidance for the teacher model to significantly strengthen audio conditioning for lip synchronization. While these strategies introduce potential instability, they represent a necessary trade-off to achieve optimal multimodal conditioning before the peak-then-degrade transition. These aggressive training strategies enable the model to learn strong audio-visual alignment (evaluated by Sync-C and Sync-D) while maintaining high visual quality.

\section{Building Real-Time Multimodal Interactive Systems}

Based on the distilled model, we build LiveTalk (Fig.~\ref{fig:system_overview}), a complete real-time multimodal interactive avatar system integrating our model with Qwen3-Omni~\cite{xu2025qwen3} for end-to-end visual communication. Our modular architecture comprises two key components: a real-time performer module (our distilled video diffusion model) that renders synchronized talking avatars, and a thinker/talker module (Qwen3-Omni) that handles reasoning and generates streaming audio responses. We detail the system pipeline overview, and the two critical system-level challenges below.

\noindent \textbf{System Pipeline Overview.} Figure~\ref{fig:system_overview} illustrates LiveTalk's interaction pipeline. When a user provides audio or text input, Qwen3-Omni~\cite{xu2025qwen3} generates streaming audio responses. Our video diffusion model takes three multimodal conditions: (1) streaming audio output $c_{\text{audio}}$ from Qwen3-Omni, (2) reference image avatar $c_{\text{img}}$ defining visual identity, and (3) text prompts $c_{\text{text}}$ describing the desired motion (e.g., "A person speaking naturally with expressive gestures and emotions"). These jointly drive block-wise autoregressive generation, where each block of $b=3$ latent frames undergoes $k = 4$ diffusion steps. Clean KV cache from previous blocks is prefilled for visual consistency, enabling streaming synchronized to audio with sub-second first-frame latency.

\noindent \textbf{Streaming Audio Conditioning.} Audio-driven video generation requires acoustic context from adjacent frames for smooth lip-sync and natural motion. Conditioning each video block solely on temporally aligned audio causes discontinuities at block boundaries. Waiting for extended audio sequences (encoding entire clips before generation) introduces prohibitive latency. Our solution uses overlapped windowing: we encode and generate as soon as a small windowed segment becomes available, providing rich acoustic context while maintaining real-time responsiveness.

\noindent \textbf{Long-Horizon Video Streaming with Speaker Identity Preservation.} Interactive dialogues often last for minutes, which requires the generated avatar's identity in the generated video to remain consistent over long time spans. Although Self Forcing~\cite{huang2025self} reduces the train–test mismatch by aligning the distributions of training and inference, identity still degrades once generation extends beyond the training window: accumulated errors can cause color drift and geometric distortions. This motivates a simple question: within a fixed attention window, can we downweight error-prone recent blocks and upweight high-fidelity identity representations instead?

Following this intuition, we propose a training-free method, \textbf{Anchor-Heavy Identity Sinks (AHIS)}: we allocate part of the KV cache as attention sinks~\cite{xiao2024efficientstreaminglanguagemodels} that permanently store early high-fidelity speaker frames as identity anchors, while the remaining rolling KV tokens maintain contextual continuity and prevent visual discontinuities. Unlike standard attention sink designs, AHIS deliberately allocates a much larger fraction of the KV window to identity sink tokens than to rolling tokens. Under the same KV-cache budget, increasing the proportion of sink tokens while reducing rolling KV tokens simultaneously strengthens the focus on high-fidelity identity and suppresses attention to accumulated errors in subsequent generations. Specifically, we set the KV window to 5 blocks, with the first three blocks used as sink tokens and the last two as rolling KV tokens. In our experiments, this setting effectively preserved the speaker’s appearance over several minutes of generated video.

\noindent \textbf{Parallel Pipeline for Video Denoising and Decoding.} Streaming video generation requires diffusion denoising (predicting clean latents) and VAE decoding (converting to pixels). Sequential execution risks playback stalling when generation time exceeds playback duration. We adopt pipeline parallelism: while the current block undergoes denoising, the previous block is simultaneously decoded. This reduces per-block latency from the sum to the maximum of the two stages, ensuring generation stays ahead of playback and enabling non-stalling streaming with seamless real-time rendering.

\section{Experiments}

\subsection{Settings}
\noindent \textbf{Model and Training Configuration.} We conduct distillation experiments on the multimodal variant of Wan2.1~\cite{wan2025wan}, instantiated by OmniAvatar~\cite{gan2025omniavatar}. OmniAvatar is a multimodal bidirectional diffusion model that accepts text, image, and audio as conditional inputs to synthesize videos with realistic facial expressions and lip synchronization.

We adapt OmniAvatar-1.3B into a causal student model by modifying its architecture to support causal attention and KV cache for AR generation. Following the block-wise AR paradigm, we adopt a block size of $b = 3$ latent frames, enabling efficient streaming generation while maintaining temporal coherence.

For training data, we curate 4000 multimodal conditions from Hallo3~\cite{cui2024hallo3} and the HDTF training split~\cite{zhang2021flow}. Note that we directly use the audio samples from the dataset, and refine the image and text conditions following the approach in Section~\ref{subsec:improvements}. Using the bidirectional OmniAvatar-1.3B, we generate corresponding ODE trajectories  from these curated conditions for ODE initialization training. 

We apply the combined distillation strategy described in Section~\ref{subsec:improvements} for on-policy DMD training. Specifically, we use OmniAvatar-14B as the frozen teacher score network $s_{\theta}$, and OmniAvatar-1.3B as the trainable critic model $s_{\psi}$. The generator is initialized via ODE trajectory matching, trained for 20k steps on the generated trajectory pairs. Subsequently, DMD distillation is performed using the same 4000 curated conditions, reaching optimal generation quality within approximately 1000 steps. Detailed hyperparameters and training configurations are provided in the Appendix.

\noindent \textbf{System Integration.} 
To build a complete real-time multimodal interactive avatar system, we integrate our distilled video generation model with Qwen3-Omni~\cite{xu2025qwen3}, which serves as the ``thinker/talker'' module responsible for reasoning and generating streaming audio responses. During interaction, the system takes the streaming audio output from Qwen3-Omni~\cite{xu2025qwen3}, along with a reference image avatar and text prompt, as conditioning inputs to our video diffusion model for real-time video generation (Fig.~\ref{fig:system_overview}).

\noindent \textbf{Evaluation Protocol.} We conduct evaluations under two settings: (1) \textbf{single-round evaluation}: standard multimodal-driven avatar video generation with a single reference image and audio clip, evaluated on established benchmarks (HDTF, AVSpeech, CelebV-HQ); and (2) \textbf{Multi-round evaluation}: free-form conversational interaction between users and the multimodal avatar system, assessed using our proposed multi-turn interaction benchmark spanning 9 evaluation dimensions.
\begin{table*}[t!]
\centering
\caption{Quantitative comparison with existing multimodal avatar generation methods on the test set. Our distilled model achieves comparable or \textit{superior} visual quality, aesthetics, and lip-sync performance to the bidirectional baseline OmniAvatar-1.3B, while delivering approximately 20× throughput speedup and 200× faster first-frame latency. \textbf{Bold} indicates best performance.}
\label{tab:face}
\tiny
\begin{tabular*}{\textwidth}{@{\extracolsep{\fill}}c|ccc|cccccc}
\Xhline{1pt}
\textbf{Methods} & \textbf{Model Size} & \textbf{Throughput (FPS)} $\uparrow$ & \textbf{Latency (s)} $\downarrow$ & \textbf{FID}   $\downarrow$ & \textbf{FVD}   $\downarrow$  & \textbf{Sync-C}  $\uparrow$  & \textbf{Sync-D}    $\downarrow$  & \textbf{IQA}  $\uparrow$  & \textbf{ASE}  $\uparrow$  \\ \hline
\multicolumn{10}{c}{HDTF} \\ \hline
Ground Truth  & - & - & - & - & - & 8.06 & 7.31 & 4.20 & 2.56  \\ 
AniPortrait \cite{wei2024aniportrait} & 2.5B & 0.38 & 211.77 & 21.41 & 323.08 & 1.17 & 13.32 & 4.03 & 2.40 \\ 
Hallo3 \cite{cui2024hallo3} & 5B & 0.21 & 589.69 & 23.18 & 290.05 & 6.01 & 9.34 & 3.54 & 2.02 \\ 
FantasyTalking \cite{wang2025fantasytalking} & 14B & 0.35 & 232.06 & 24.54 & 488.53 & 2.92 & 12.21 & 3.78 & 2.14 \\ 
OmniAvatar-14B \cite{gan2025omniavatar} & 14B & 0.20 & 412.50 & \textbf{8.82} & \textbf{151.30} & \textbf{6.29} & \textbf{9.16} & 3.91 & 2.30 \\
OmniAvatar-1.3B \cite{gan2025omniavatar} & 1.3B & 0.97 & 83.44 & 10.85 & 187.46 & 3.85 & 11.38 & 3.87 & 2.25 \\
Ours & 1.3B & \textbf{24.82} & \textbf{0.33} & 13.68 & 190.07 & 4.50 & 11.00 & \textbf{4.13} & \textbf{2.52} \\
\hline
\multicolumn{10}{c}{AVSpeech} \\ \hline
Ground Truth  & - & - & - & - & - & 5.83 & 8.19 & 4.11 & 2.50 \\ 
AniPortrait \cite{wei2024aniportrait} & 2.5B & 0.38 & 211.77 & 32.14 & 553.27 & 0.84 & 12.86 & 4.07 & 2.49  \\ 
Hallo3 \cite{cui2024hallo3} & 5B & 0.21 & 589.69 & 32.88 & 502.05 & 5.10 & 9.50 & 3.49 & 1.99  \\ 
FantasyTalking \cite{wang2025fantasytalking} & 14B & 0.35 & 232.06 & 33.08 & 509.56 & 2.16 & 11.72 & 3.68 & 2.15  \\ 
OmniAvatar-14B \cite{gan2025omniavatar}& 14B & 0.20 & 412.50 & \textbf{31.21} & \textbf{450.09} & \textbf{5.18} & \textbf{9.17} & 3.77 & 2.25 \\
OmniAvatar-1.3B \cite{gan2025omniavatar} & 1.3B & 0.97 & 83.44 & 31.55 & 483.94 & 3.16 & 10.91 & 3.70 & 2.18 \\
Ours & 1.3B & \textbf{24.82} & \textbf{0.33} & 33.96 & 486.77 & 3.71 & 10.79 & \textbf{4.08} & \textbf{2.50} \\
\hline
\multicolumn{10}{c}{CelebV-HQ} \\ \hline
Ground Truth  & - & - & - & - & - & 5.36 & 8.13 & 4.31 & 2.82  \\ 
AniPortrait \cite{wei2024aniportrait} & 2.5B & 0.38 & 211.77 & 23.87 & 427.43 & 1.14 & 12.07 & 4.24 & 2.79  \\ 
Hallo3 \cite{cui2024hallo3} & 5B & 0.21 & 589.69 & 27.15 & 432.18 & 4.65 & 9.20 & 3.77 & 2.35  \\ 
FantasyTalking \cite{wang2025fantasytalking} & 14B & 0.35 & 232.06 & 26.87 & 454.98 & 2.48 & 10.95 & 3.93 & 2.46  \\ 
OmniAvatar-14B \cite{gan2025omniavatar} & 14B & 0.20 & 412.50 & \textbf{21.25} & 406.78 & \textbf{4.81} & \textbf{8.85} & 4.03 & 2.59 \\
OmniAvatar-1.3B \cite{gan2025omniavatar} & 1.3B & 0.97 & 83.44 & 22.28 & \textbf{382.35} & 3.09 & 10.40 & 3.98 & 2.52 \\

Ours & 1.3B & \textbf{24.82} & \textbf{0.33} & 25.37 & 437.97 & 3.78 & 10.08 & \textbf{4.29} & \textbf{2.79} \\
\Xhline{1pt}
\end{tabular*}
\end{table*}

\subsection{Single-Round Evaluation}
We evaluate our distilled model against several baselines including AniPortrait~\cite{wei2024aniportrait}, Hallo3~\cite{cui2024hallo3}, FantasyTalking~\cite{wang2025fantasytalking}, and the bidirectional teachers OmniAvatar-1.3B and OmniAvatar-14B~\cite{gan2025omniavatar}. The evaluation is conducted on 100 randomly sampled 5-second clips across three multimodal-driven avatar video generation benchmarks: HDTF test split~\cite{zhang2021flow} for in-domain evaluation, and AVSpeech~\cite{ephrat2018looking} and CelebV-HQ~\cite{zhu2022celebv} for out-of-domain evaluation. Following the evaluation protocol in~\cite{gan2025omniavatar, chen2025hunyuanvideo}, we employ FID~\cite{fid}, FVD~\cite{fvd}, IQA~\cite{qalign}, and ASE~\cite{qalign} to assess visual quality and aesthetics, and Sync-C/D~\cite{syncnet} to measure lip-sync synchronization between the audio condition and the generated lip movements. Ground truth videos are also evaluated on reference-free metrics for comparison. To assess inference efficiency, we measure the throughput and first-frame latency of all models at $512\times512$ resolution on a single GPU. The evaluation protocol consists of several warm-up generations followed by multiple test runs with randomly sampled conditions, with all metrics averaged across the test runs to ensure statistical reliability.

\begin{figure}[htbp]
    \centering
    \includegraphics[page=1, width=0.7\textwidth]{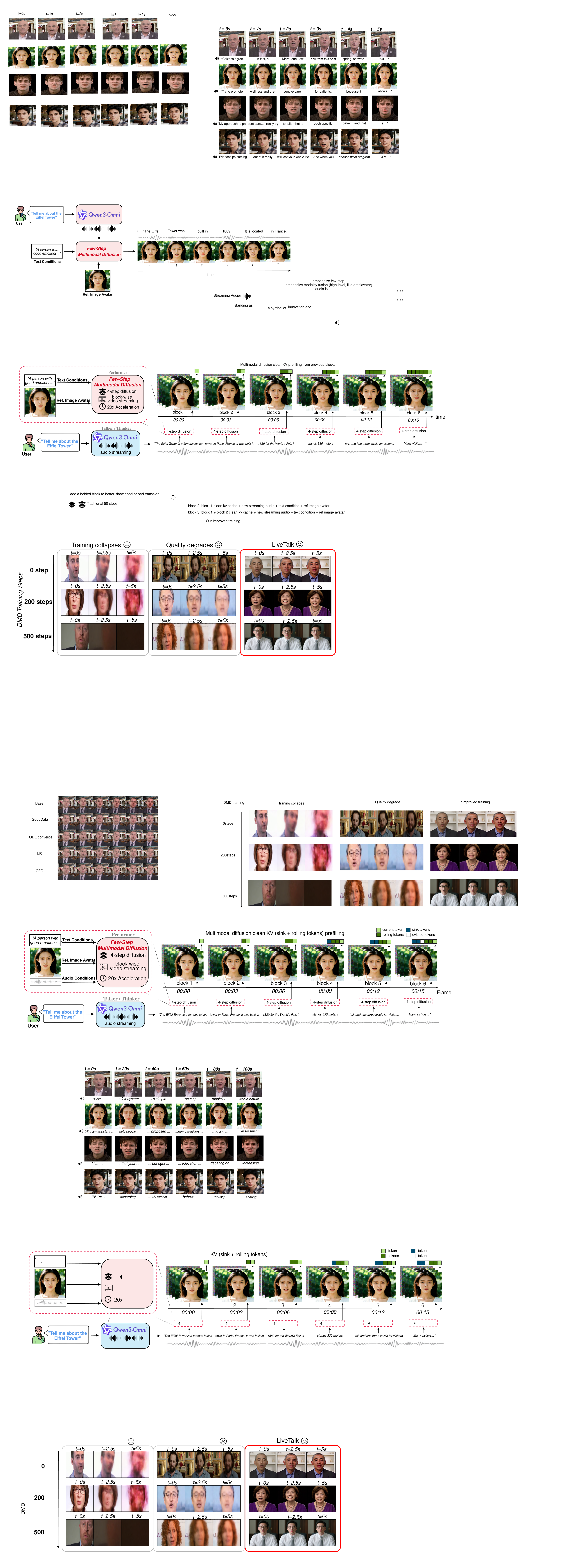}
    \vspace{-2mm}
    \caption{\textbf{Examples of multimodal-conditioned avatar video generation by our model.} Our model generates temporally coherent video with natural facial expressions, accurate lip-sync to the audio conditions, and consistent visual identity across frames.}
    \vspace{-4mm}
    \label{fig:demo}
\end{figure}

\noindent \textbf{Results and Analysis.} 
Our distilled model achieves comparable or \textit{superior} visual quality, aesthetics, and lip-sync synchronization compared to the bidirectional many-step variant (i.e., OmniAvatar-1.3B~\cite{gan2025omniavatar}) across in-domain HDTF~\cite{zhang2021flow} and out-of-domain AVSpeech~\cite{ephrat2018looking} and CelebV-HQ benchmarks. More significantly, our model demonstrates substantial efficiency gains: 24.82 FPS throughput compared to 0.97 FPS (25× speedup) and first-frame latency reduced from 83.44s to 0.33s (250× faster). Remarkably, our 1.3B distilled model achieves comparable or superior performance to several larger bidirectional, many-step models, including AniPortrait (2.5B), Hallo3 (5B), and FantasyTalking (14B), while maintaining significantly higher efficiency (over 100× improvement in latency and 50× in throughput). Figure~\ref{fig:demo} shows representative video samples generated by our model.

\subsection{Multi-Round Evaluation}
Existing evaluation benchmarks for multimodal-driven avatar video generation primarily focus on single-audio-clip metrics, using traditional visual evaluation measures such as lip-sync accuracy (Sync-C/D), image quality (FID/FVD/IQA), and aesthetics (ASE). However, these metrics fail to assess multi-turn interaction quality, a critical requirement for real-world multimodal conversational applications. To address this gap, we propose a multi-round interaction benchmark and an evaluation protocol based on vision-language models (VLMs)~\cite{Qwen-VL, lee2024prometheus} to comprehensively evaluate the multi-turn conversational capabilities of audio-driven avatar systems.

\noindent \textbf{Benchmark Design Methodology.} We meticulously curate 100 multi-turn evaluation scenarios that require multimodal interactive AI systems to provide coherent audio-driven video generation responses across conversational turns. For example, in the first round, a user might ask, ``Tell me about the Eiffel Tower,'' and the system should generate a coherent introduction to the landmark with synchronized audio and video. In the subsequent round, the user might follow up with, ``Where is this avenue located?'' referencing the Champs-Élysées mentioned in the previous response. The AI system must generate coherent video and audio outputs that are contextually grounded in its previous multimodal conversational history, demonstrating the ability to maintain temporal coherence and contextual awareness across multiple interaction turns.

\noindent \textbf{Evaluation Protocol.} We adopt Qwen3-VL-30B-A3B-Instruct~\cite{bai2025qwen2} as the VLM-as-evaluator~\cite{lee2024prometheus} with structured prompts tailored to each dimension. Our evaluation framework comprises two primary categories: \textit{Visual Interaction Performance} (4 dimensions: Emotional Appropriateness, Nonverbal Interaction, Multi-Video Coherence, and Conversational Naturalness) and \textit{Interaction Content Quality} (5 dimensions: Semantic Relevance, Information Completeness, Logical Consistency, Context Understanding, and Overall Interaction Experience). Detailed evaluation dimension descriptions are provided in the Appendix. System-generated audio responses are transcribed via FunASR~\cite{gao2023funasr} to enable audio-video synchronization and content quality evaluation. 

To enable fair comparison across baselines, we normalize raw scores for each metric using z-score transformation. Specifically, for each metric, we create a single pooled distribution containing scores from all methods on all evaluation samples, compute z-scores as $z = (x - \mu) / \sigma$ where $\mu$ and $\sigma$ are from the pooled distribution, and convert each z-score to its percentile (0-100). We report average percentiles per baseline for each metric, providing relative performance interpretation.

\begin{table*}[htp!]
\centering
\scriptsize
\caption{Multi-Round Interaction Quality Evaluation. Our method is benchmarked against baselines on the proposed interaction benchmark. We report Z-Score percentile values for Visual Interaction Performance and Interaction Content Quality metrics, along with average inference throughput and latency per turn. \textbf{Bold} indicates best performance, and \underline{underline} indicates second-best.}
\label{tab:interaction_benchmark_method_a}
\scriptsize
\begin{tabular*}{\textwidth}{@{\extracolsep{\fill}}lccccccccccc}
\toprule
\multirow{2}{*}{\textbf{Method}}
& \multirow{2}{*}{\textbf{Throughput (FPS)} $\uparrow$}
& \multirow{2}{*}{\textbf{Latency (s)} $\downarrow$}
& \multicolumn{4}{c}{\textbf{Visual Interaction Performance} $\uparrow$}
& \multicolumn{5}{c}{\textbf{Interaction Content Quality} $\uparrow$} \\
\cmidrule(lr){4-7} \cmidrule(lr){8-12}
& &
& EA & NI & MVC & CN
& SR & IC & LC & CU & OIE \\
\midrule
Veo3
& - & 61.46
& 23.51
& 24.68
& \underline{26.68}
& 24.93
& 25.17
& 17.21
& 20.56
& 20.05
& 19.07
\\
Sora2
& - & 121.85
& \textbf{75.78}
& \textbf{72.20}
& 25.85
& \textbf{74.32}
& \underline{66.68}
& \underline{50.01}
& \underline{65.07}
& \underline{68.80}
& \underline{53.09}
\\
\textbf{LiveTalk}
& 24.82 & 1.16
& \underline{59.54}
& \underline{60.42}
& \textbf{87.26}
& \underline{56.32}
& \textbf{72.02}
& \textbf{81.27}
& \textbf{75.20}
& \textbf{72.65}
& \textbf{81.59}
\\
\bottomrule
\end{tabular*}

\vspace{0.5em}
\footnotesize
\textit{Note:} EA: Emotional Appropriateness, NI: Nonverbal Interaction, MVC: Multi-Video Coherence, CN: Conversational Naturalness;
SR: Semantic Relevance, IC: Information Completeness, LC: Logical Consistency, CU: Context Understanding, OIE: Overall Interaction Experience.
\end{table*}

\begin{table*}[htbp]
\centering
\scriptsize
\caption{Ablation study showing the impact of various improvements. Each row sequentially adds one component: (1) curated high-quality multimodal conditions, (2) training ODE distillation to convergence, (3) accelerated learning rate, and (4) tuned CFG scale for real-score estimation. "Final Configuration" is the best performing configuration after all improvements are applied. The last row represents the baseline multimodal conditions with all other improvements applied.}
\label{tab:ablation-four-factors}
\footnotesize
\begin{tabular*}{\textwidth}{@{\extracolsep{\fill}}lcccccc}
\toprule
Setting & \textbf{FID}$\downarrow$ & \textbf{FVD}$\downarrow$ & \textbf{Sync-C}$\uparrow$ & \textbf{Sync-D}$\downarrow$ & \textbf{IQA}$\uparrow$ & \textbf{ASE} \\
\midrule
Baseline & 27.10 & 338.08  & 3.13 & 11.77 &3.95  & 2.38 \\
+ Curated Multimodal Conditions &  14.90&  217.68&3.53 & 11.47 & 3.99 &  2.31\\
+ Converged ODE Initialization & \textbf{11.67} &  \textbf{169.75}&  4.15&  11.19& \textbf{4.18} & \textbf{2.56} \\
+ Aggressive Learning Rate & 12.10 & 179.73 & 4.29 & 11.07 & 4.15 &  2.53\\
+ Tuned Teacher Score CFG \textbf{(Final Configuration)} & 13.68 & 190.07 & \textbf{4.50} & \textbf{11.00} & 4.13 & 2.52 \\
\midrule
Final Configuration \textbf{without} Curated Multimodal Conditions & 23.89 & 261.19 & 3.85 & 11.42 & 4.06 & 2.47 \\
\bottomrule
\end{tabular*}
\end{table*}

\noindent \textbf{Results and Analysis.} 
Table~\ref{tab:interaction_benchmark_method_a} presents multi-round evaluation against Veo3~\cite{deepmind_veo3_2025} and Sora2~\cite{openai_sora2_2025}. LiveTalk outperforms both baselines in multi-video coherence and content quality metrics while remaining competitive on other visual interaction dimensions, demonstrating its advantages in maintaining meaningful and coherent multi-turn multimodal interaction. Notably, both Veo3 and Sora2 exhibit visual drift across turns due to prolonged generation times (61-122s) that break conversational flow and lack of effective memory mechanisms for multi-round video interaction. In contrast, LiveTalk adopts AR generation with KV cache to maintain visual memory states, while also leveraging the Qwen3-Omni thinker/talker module to preserve textual memory states, enabling coherent generation across  modalities while preserving real-time responsiveness.

\subsection{Ablations}
\label{subsec:ablations}

We conduct ablation studies on the four proposed components in Section~\ref{subsec:improvements}: (1) refined multimodal conditions, (2) converged ODE initialization, (3) aggressive learning rate schedule, and (4) tuned teacher score CFG guidance. We evaluate on the HDTF test set using 100 randomly sampled 5-second clips, measuring both visual quality metrics (FID, FVD, IQA, ASE) and audio-visual synchronization metrics (Sync-C, Sync-D). Table~\ref{tab:ablation-four-factors} presents the results, with each row sequentially adding one component to demonstrate its incremental contribution.

\begin{figure}[h]
    \centering
    \includegraphics[page=1, width=0.6\textwidth]{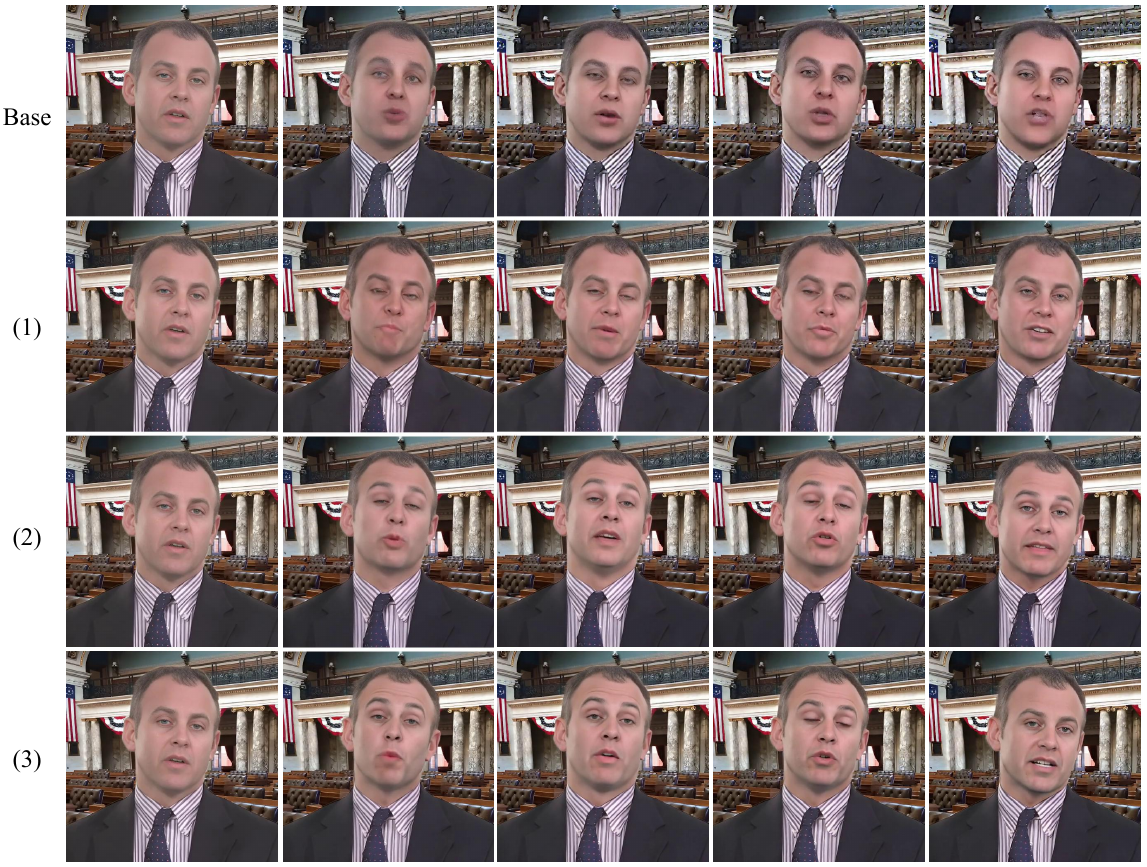}
    \vspace{-2mm}
    \caption{Ablation study visualization. Generated video shows progressive improvements for each ablated component: (1) curated multimodal conditions, (2) ODE initialization for full convergence (20k steps), and (3) aggressive hyperparameter settings (doubled learning rates, CFG=6).}
    \vspace{-4mm}
    \label{fig:ablation_visual}
\end{figure}

\noindent \textbf{Results and Analysis.} We systematically evaluate the impact of each design choice on distillation quality.

\textbf{Baseline.} Our baseline uses a combination of 4000 multimodal conditions (2000 from Hallo3, and 2000 from HDTF), selected based on brightness and image quality metrics of the reference image. We then follow default Self Forcing distillation settings: ODE initialization for 4000 steps, followed by 1000 DMD steps (critic learning rate $4\times10^{-7}$, generator learning rate $2\times10^{-6}$, teacher score CFG scale 4) with EMA decay 0.99. This configuration has the poorest performance, with generated videos exhibiting severe quality degradation.

\textbf{Curated Multimodal Conditions.} Training on curated multimodal conditions, the distilled model generates videos with substantially improved visual quality and audio-visual synchronization. While severe degradation is eliminated, inter-frame artifacts including blurriness and flickering remain noticeable. 

\textbf{Converged ODE Initialization.} Extending ODE initialization to 20000 steps for full convergence, the model completely resolves these visual defects, with all quality metrics reaching peak performance at this stage.

\textbf{Aggressive LR \& Tuned Teacher Score CFG.} Finally, we explore a more aggressive hyperparameter setting by doubling both learning rates and increasing the teacher score CFG scale to 6. These modifications yield substantial improvements in lip-sync accuracy at a modest cost to visual quality, with mouth movements becoming significantly more pronounced and articulated. However, we find that further increasing these parameters results in either training instability or visual oversaturation.

\textbf{Final Configuration without Curated Multimodal Conditions.} To verify the importance of curated conditions, we apply our other improvements to the baseline's multimodal conditions. Results show that visual quality remains degraded when distilling with lower-quality conditions, with persistent issues including poor temporal consistency and abrupt color shifts between frames, highlighting that data curation is essential for successful distillation.

Examples of generated videos from each configuration are shown in Figure~\ref{fig:ablation_visual}. These results validate the effectiveness of each of our proposed improvement in enhancing generation quality and stabilizing multimodal distillation.

\section{Conclusion}
We propose an improved on-policy distillation recipe that distills a bidirectional, many-step video diffusion model into a causal, 4-step AR video diffusion. Our recipe incorporates refined multimodal conditioning, converged ODE initialization, and aggressive optimization to achieve strong visual fidelity and audio-visual synchronization while delivering sub-second first-frame latency and real-time throughput. Building on the distilled model, we develop LiveTalk, a real-time multimodal interactive avatar system that streams video conditioned on text, image, and audio. LiveTalk delivers coherent multi-turn interactions with significantly lower latency than state-of-the-art video models. These results enable new possibilities for real-time multimodal human-AI interactive systems.

\clearpage

\bibliographystyle{acl_natbib}
\bibliography{bib}
\appendix
\clearpage

\section{Multi-Round Evaluation Details}
\label{sec:appendix_prompts}

This appendix provides a comprehensive description of the nine evaluation dimensions comprising our interaction benchmark. Each dimension is assessed by a VLM using carefully designed, structured prompts that define specific scoring criteria. The evaluation framework is divided into two major categories: \textbf{Visual Interaction Performance} and \textbf{Interaction Content Quality}. We also provide extra implementation details for the evaluation protocol.

\subsection{Visual Interaction Performance}

Visual interaction performance captures the non-verbal and emotional aspects of the assistant's video responses, emphasizing the quality of human-like engagement and visual consistency across conversational turns.

\noindent \textbf{Emotional Appropriateness.}
This dimension evaluates the degree to which the assistant's facial expressions align with conversational context and emotional content.

Assessment criteria include:
\begin{itemize}[leftmargin=*,noitemsep,topsep=0pt]
    \item \textbf{Content-Emotion Matching:} Appropriateness of emotional expressions relative to semantic content (e.g., conveying seriousness during complex explanations; displaying warmth during positive encouragement).
    \item \textbf{Intensity Calibration:} Proportionality of emotional intensity to the significance and tone of the conversational content.
    \item \textbf{Emotional Transitions:} Naturalness and coherence of emotional shifts across multiple consecutive video responses.
    \item \textbf{User Responsiveness:} Adaptive emotional responses that acknowledge and appropriately react to the user's inferred emotional state (e.g., providing reassurance for confusion, matching enthusiasm).
\end{itemize}

\noindent \textbf{Nonverbal Interaction.}
This dimension quantifies the quality of interactive nonverbal cues that establish and maintain user engagement. 

Assessment criteria include:
\begin{itemize}[leftmargin=*,noitemsep,topsep=0pt]
    \item \textbf{Eye Contact Quality:} Maintenance of natural, engaging gaze patterns that simulate direct interaction with the user.
    \item \textbf{Interactive Gestures \& Micro-expressions:} Effective use of confirmatory nods, thoughtful pauses, raised eyebrows, and other subtle movements that signal active participation in dialogue.
    \item \textbf{Listening Response:} Presence of nonverbal acknowledgment signals that indicate comprehension and processing of user input prior to verbal response.
   \item \textbf{Interaction Authenticity:} Overall impression of genuine bidirectional dialogue rather than unidirectional content delivery.
\end{itemize}

\noindent \textbf{Multi-Video Coherence.}
This dimension assesses visual and behavioral consistency across multiple video responses within the conversation. This dimension is important as it evaluates the immersion and believability in multi-round interactions, which is especially important in extended interactions. 

Assessment criteria include:
\begin{itemize}[leftmargin=*,noitemsep,topsep=0pt]
    \item \textbf{Visual Identity Consistency:} Stability of the speaker's physical appearance, including facial features, hairstyle, and attire.
    \item \textbf{Scene Consistency:} Stability of environmental factors including background elements, lighting conditions, and camera perspective.
    \item \textbf{Emotional Continuity:} Logical progression and smooth evolution of emotional states from one video response to subsequent responses.
    \item \textbf{Behavioral Coherence:} Consistency in posture, gestural patterns, speech cadence, and overall presentation style.
\end{itemize}

\noindent \textbf{Conversational Naturalness.}
This dimension measures the degree to which the interaction conveys authentic human-like spontaneity rather than scripted or mechanistic behavior. 

Assessment criteria include:
\begin{itemize}[leftmargin=*,noitemsep,topsep=0pt]
    \item \textbf{Dialogue vs. Broadcast Distinction:} Presence of interactive, bidirectional conversational dynamics versus formal, unidirectional presentation style.
    \item \textbf{Spontaneous Human Characteristics:} Natural occurrences of brief pauses, thinking moments, speech hesitations, and other markers of authentic spontaneous communication.
\end{itemize}

\subsection{Interaction Content Quality}

Interaction content quality evaluates the semantic, logical, and contextual appropriateness of the assistant's verbal responses, ensuring both accuracy and conversational coherence.

\noindent \textbf{Semantic Relevance.} This dimension assesses whether the assistant's responses directly and accurately address the topical focus of user queries. 

\noindent \textbf{Information Completeness.} This dimension evaluates if the provided answers are comprehensive, sufficiently detailed, and self-contained. 

\noindent \textbf{Logical Consistency.} This dimension identifies contradictions, factual errors, or logical inconsistencies both within individual responses and across the entire conversational history. 

\noindent \textbf{Context Understanding.} This dimension measures the assistant's ability to maintain conversational state, correctly resolve anaphoric references, and effectively build upon information established in prior dialogue turns. 

\noindent \textbf{Overall Interaction Experience.} This dimension provides a holistic assessment encompassing conversational fluency, effectiveness in task completion, and overall user satisfaction. 

\subsection{Evaluation Protocol Details}
\label{sec:appendix_impl}

\noindent \textbf{VLM Evaluator.} We employ Qwen2.5-VL-30B-A3B-Instruct~\cite{Qwen2.5-VL} as our evaluation model, deployed via vLLM~\cite{kwon2023efficient}. The model configuration supports contexts up to 256K tokens and enhanced video understanding capabilities, processing up to 32 frames per video at 784$\times$784 resolution. To ensure stable and consistent scoring, we use a low sampling temperature of 0.1 combined with top-p sampling (p=0.9). Each evaluation dimension is guided by a detailed rubric encoded in dimension-specific prompt files, with model responses structured as \texttt{\textbackslash boxed\{score\}} for automated parsing and aggregation.


\noindent \textbf{ASR Transcription.} Full conversational transcripts are constructed by transcribing both user audio inputs (16kHz WAV/MP3 format) and assistant video audio tracks into English text. We utilize the FunASR (\texttt{speech-\allowbreak{}paraformer\allowbreak{}-asr-\allowbreak{}en-\allowbreak{}16k-\allowbreak{}vocab4199})~\cite{gao2023funasr} model, which employs a non-autoregressive Paraformer architecture optimized for high-accuracy real-time transcription.


\noindent \textbf{Multi-Video Evaluation.} 
For conversations containing multiple assistant video responses, all video files are presented to the VLM evaluator in chronological order alongside the complete conversational transcript. This multi-video presentation enables both isolated analysis of individual video segments and comparative cross-video analysis, which is essential for accurately assessing the \textit{Multi-Video Coherence} dimension and evaluating \textit{Emotional Transitions}. All content quality dimensions leverage the full conversation history to assess semantic coherence, logical consistency, and contextual understanding across all dialogue turns.

\newpage
\section{Training Configuration Details}

The distillation pipeline consists of two sequential stages: ODE initialization and on-policy distribution matching distillation (DMD) (Section 3.1). Tab.~\ref{tab:training_config} details the hyperparameters and optimization settings for both stages.

\begin{table}[htbp]
\caption{\textbf{Training configurations for ODE trajectory initialization and DMD distillation.} CFG (Generating Trajectory) denotes the classifier-free guidance scale used by the bidirectional model (OmniAvatar-1.3B) during ODE trajectory generation. For DMD distillation, the critic score network $s_{\psi}$ is first updated for 20 steps to obtain accurate score estimates, after which the student generator network $g_{\phi}$ is trained (Section~3.1). Exponential moving average (EMA) updates are enabled from step 200 onward.}
\label{tab:training_config}
\centering
\small
\begin{tabular}{l c}
\toprule

\textbf{Hyperparameters}& \textbf{ODE Initialization}   \\ 
\midrule
Batch size & 64 \\
 Learning rate & $4\mathrm{e}{-5}$ \\
 Optimizer  &
\begin{tabular}[c]{@{}c@{}}
AdamW, $\beta_1 = 0.9$, $\beta_2 = 0.999$,\\
$\epsilon = 1\mathrm{e}{-8}$,weight\_decay=0
\end{tabular}
\\
CFG (Generating Trajectory)&4.5\\
 \midrule
 \textbf{Hyperparameters} & \textbf{DMD Distillation} \\
\midrule
Teacher score network & OmniAvatar-14B \\
Teacher score CFG  & 6.0 \\
Critic score network  & OmniAvatar-1.3B \\
Batch size & 64 \\
Optimizer ($g_\phi$) &
\begin{tabular}[c]{@{}c@{}}
AdamW, $\beta_1 = 0$, $\beta_2 = 0.999$,\\
$\epsilon = 1\mathrm{e}{-8}$, weight\_decay=0.01
\end{tabular}
\\
Optimizer ($s_\psi$) &
\begin{tabular}[c]{@{}c@{}}
AdamW, $\beta_1 = 0$, $\beta_2 = 0.999$,\\
$\epsilon = 1\mathrm{e}{-8}$, weight\_decay=0.01
\end{tabular}
\\
Learning rate ($g_\phi$) & $4\mathrm{e}{-6}$ \\
Learning rate ($s_\psi$) & $8\mathrm{e}{-7}$ \\
Generator/critic update ratio & 5, with critic warmup for 20 steps \\
EMA decay & 0.99 \\

\bottomrule

\end{tabular}
\end{table}

\end{document}